%% file: paper.tex
\documentclass[11pt,a4paper]{article}
\usepackage[hyperref]{emnlp2020}
\usepackage{times}
\usepackage{latexsym}

\usepackage{booktabs} 
\usepackage{color}
\usepackage{hyperref}
\usepackage{url}
\usepackage{amsmath}
\usepackage{amsfonts}
\usepackage{graphicx}
\usepackage{bbm}
\usepackage{multirow, multicol}
\usepackage{comment}
\usepackage{xspace}
\usepackage{upquote}
\usepackage{float}
\usepackage{alltt}
\usepackage{algpseudocode}
\usepackage{pbox}
\usepackage{enumitem}
\usepackage{mathpartir}
\usepackage{wrapfig}
\usepackage{array}
\usepackage{hhline}
\usepackage{stmaryrd}
\usepackage{pifont}
\usepackage{makecell}
\usepackage{cuted}
\usepackage{subcaption}

\usepackage{microtype}

\aclfinalcopy 
\def\aclpaperid{3506} 
\definecolor{comment-red}{rgb}{0.8,0,0}
\newcommand{\TODO}[1]{{\color{blue} TODO: #1}}

\newcommand{\SQA}[0]{Schema2QA\xspace}

\newcommand{\webqa}[0]{Schema2QA\xspace}

\newcommand{\system}[0]{AutoQA\xspace}
\newcommand{\overnight}[0]{{Overnight}\xspace}
\newcolumntype{P}[1]{>{\centering\arraybackslash}p{#1}}

\newcommand{\tttable}[0]{{\color{blue} table}\xspace}
\newcommand{\ttvalue}[0]{{\color{blue} value}\xspace}
\title{AutoQA: From Databases To QA Semantic Parsers\\
With Only Synthetic Training Data}

\makeatletter
\newcommand{\printfnsymbol}[1]{%
  \textsuperscript{\@fnsymbol{#1}}%
}
\makeatother
  
\author{Silei Xu\thanks{\hspace{1em}Equal contribution} \quad Sina J. Semnani\printfnsymbol{1} \quad Giovanni Campagna \quad Monica S. Lam \\
  Computer Science Department \\
  Stanford University \\
  Stanford, CA, USA \\
  \texttt{\{silei,sinaj,gcampagn,lam\}@cs.stanford.edu}}
\hypersetup{pdfauthor={Silei Xu, Sina J. Semnani, Giovanni Campagna, Monica S. Lam}}

\date{}

\begin{document}
\maketitle

\input{abstract}

\input{intro}

\input{related}

\input{local-paraphrase}

\input{global-paraphrase}
\input{experiments}
\input{conclusion}

\section*{Acknowledgements}
This work is supported in part by the National Science Foundation
under Grant No.~1900638 and the Alfred P. Sloan Foundation under Grant No.~G-2020-13938.

\bibliographystyle{acl_natbib}
\bibliography{paper}

\input{appendix}
\end{document}

%% file: abstract.tex
\begin{abstract}


We propose \system, a methodology and toolkit to generate semantic parsers that answer questions on databases, with no manual effort. 
Given a database schema and its data, AutoQA automatically generates a large set of high-quality questions for training that covers
different database operations.
It uses automatic paraphrasing combined with template-based parsing to find alternative expressions of an attribute in different parts of speech. It also uses a novel filtered auto-paraphraser to generate correct paraphrases of entire sentences.

We apply \system to the \webqa dataset and obtain an average logical form accuracy 
of 62.9\% when tested on natural questions, which is only 6.4\% lower than a model trained with expert natural language annotations and paraphrase data collected from crowdworkers.  To demonstrate the generality of AutoQA, we also apply it to the \overnight dataset. \system achieves 69.8\% answer accuracy, 16.4\% higher than the state-of-the-art zero-shot models and only 
5.2\% lower than the same model trained with human data.
\end{abstract}


%% file: intro.tex
\section{Introduction}
Semantic parsing is the task of mapping natural language sentences to executable logical forms.
It has received significant attention in question answering systems for structured data ~\cite{overnight, zhong2017seq2sql, DBLP:journals/corr/abs-1809-08887, xu2020schema2qa}. 
However, training a semantic parser with good accuracy requires a large amount of annotated data, which is expensive to acquire. The complexity of logical forms means annotating the data has to be done by an expert. This adds to the cost and hinders extending question answering to new databases and domains.

\begin{figure}[htb]
\centering
\includegraphics[width=1\linewidth]{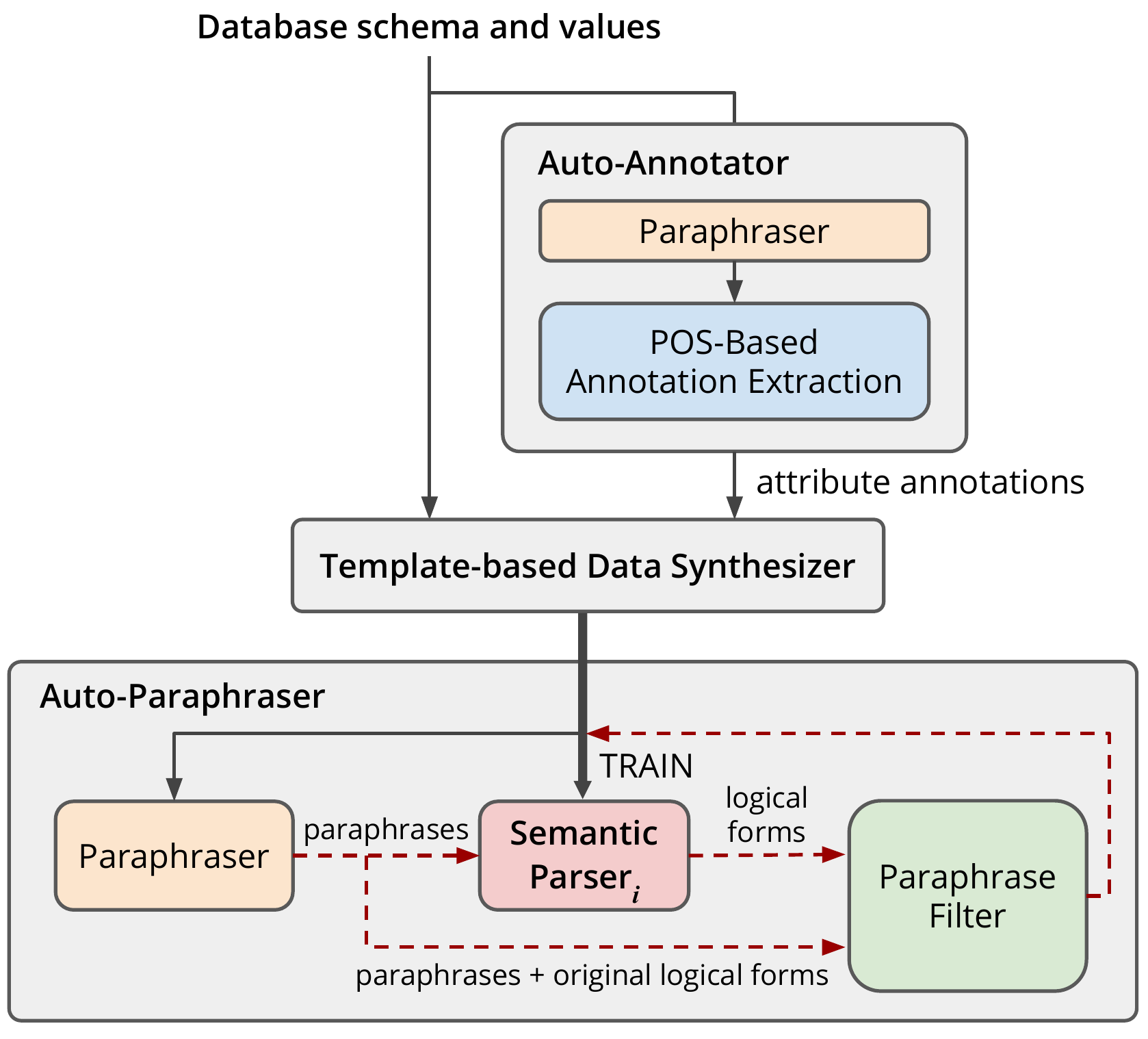}
\caption{The architecture of the \system toolkit. (a) The auto-annotator extracts annotations from paraphrases. (b) A template-based data synthesizer~\cite{xu2020schema2qa} generates data from the annotations to train a semantic parser. (c) An auto-paraphraser uses self-training to iteratively introduce more paraphrases to train the next version of the semantic parser. The red dotted lines show that generated paraphrases are filtered out unless the current semantic parser can translate them to the logical forms of the original sentences.}
\label{fig:diagram}
\end{figure}

To eliminate the need for annotating data with logical forms, SEMPRE~\cite{overnight} proposed the new methodology of first synthesizing questions on the database, then manually paraphrasing them. Recently, the Schema2QA toolkit ~\cite{xu2020schema2qa} demonstrated that it is possible to achieve high accuracy on realistic user inputs using this methodology with a comprehensive set of generic, domain-independent question templates. However, this approach requires a significant manual effort for each domain: the developers must supply how each attribute can be referred to using different parts of speech, and crowdworkers are needed to paraphrase the queries. 

Our objective is to eliminate the need for manual effort in building semantic parsers, while achieving comparable accuracy.
We hypothesize that, for common domains, the knowledge of how each attribute would be referred to in natural language is implicitly presented in large text corpora and can be captured by general-purpose paraphrasing models. With that insight, we developed AutoQA, a toolkit that (1) automatically annotates the database attributes using paraphrasing models, (2) uses generic templates to synthesize a large set of complex queries, and (3) uses a novel filtered auto-paraphraser to further increase the variety of the synthesized data. The resulting dataset is then used to train a BERT-LSTM model~\cite{xu2020schema2qa}. The architecture of AutoQA is shown in Fig.~\ref{fig:diagram}.


The contributions of this paper are:
\begin{itemize}
\item \system, a toolkit that automatically creates a semantic parser that answers questions about a given database. As the parser is trained only with automatically generated data, its cost is significantly lower than current approaches. 

\item 
A novel algorithm for annotating database attributes with phrases in different parts of speech. The algorithm is based on automatic paraphrasing combined with template-based parsing (Section \ref{sec:auto-annotation}).

\item A new automatic paraphrasing model, based on BART~\cite{bart}, that can generate natural paraphrases of sentences, with a filter trained with synthetic data to ensure the preservation of the original meaning expressed in a formal language (Section \ref{sec:autoparaphase}).

\item The methodology has been tested on the 
\overnight dataset~\cite{overnight} and Schema.org web data~\cite{xu2020schema2qa} (Section \ref{sec:experiments}).
On \overnight, \system achieves an average of 55.6\% logical form accuracy and 69.8\% denotation (answer) accuracy without using the human paraphrases for training, which are 18.6\% and 16.4\% higher than the state-of-the-art zero-shot models, respectively.
On Schema.org, \system achieves an average logical form accuracy
of 62.9\%, within 6.4\% of models trained with manual annotations and human paraphrases.\footnote{The data and code can be downloaded from \url{https://stanford-oval.github.io/schema2qa}}
\end{itemize}

%% file: related.tex
\section{Related Work}
\label{sec:related}



\paragraph{Bootstrapping Semantic Parsers.}
Neural semantic parsing for question answering is a well-known research topic~\cite{pasupat2015compositional, overnight, dong2016language, jia2016data, krishnamurthy2017neural, zhong2017seq2sql, DBLP:journals/corr/abs-1809-08887}. State of the art methods use a sequence-to-sequence architecture with attention and copying mechanism~\cite{dong2016language, jia2016data} and rely on large datasets.
Acquiring such datasets is expensive, and the work must be replicated in every new domain. 

Prior work proposed bootstrapping semantic parsers using paraphrasing~\cite{overnight}, where a dataset is synthesized using a grammar of natural language, and then paraphrased by crowdworkers to form the training set. Paraphrasing has been applied to datasets for SQL~\cite{zhong2017seq2sql}, as well as multi-turn dialogue datasets~\cite{shaw2018self, rastogi2019towards}.

Our previous work with Genie \cite{geniepldi19} proposed training with large amounts of synthesized and smaller amounts of paraphrased data. Later, we developed Schema2QA \cite{xu2020schema2qa}, a synthesis tool based on a general grammar of English. Schema2QA was found to be effective for the question answering task on the Web. Both works rely on manual paraphrases and hand-tuned annotations on each database attribute. 
Training with synthetic data has also been explored to complement existed dataset~\cite{DBPal} and in the few-shot setting~\cite{zeroshotmultiwoz, spl}.

\begin{table*}
\small
\centering
\begin{tabular}{l}
\toprule
{\bf Question}: Show me 5-star restaurants with more than 100 reviews? \\
{\bf ThingTalk}: Restaurant, aggregateRating.ratingValue == 5 \&\& aggregateRating.reviewCount \textgreater= 100\\
\midrule
{\bf Question}: What's the phone number of the McDonald's on Parker Road? \\ 
{\bf ThingTalk}: [telephone] of (Restaurant, id =~ ``McDonald's'' \&\& geo == new Location(``Parker Road'')\\
\midrule
{\bf Question}: Which is the best Chinese restaurants around here? \\
{\bf ThingTalk}: sort aggregateRating.ratingValue desc of (Restaurant, geo == HERE \&\& servesCuisine =$\sim$ ``Chinese'')\\
\bottomrule
\end{tabular}
\caption{Example questions in the restaurant domain with their ThingTalk representations.}
\label{table:thingtalk}
\end{table*}

A different line of work proposed training with a large multi-domain dataset, and then using transfer learning to generalize to new datasets, in a completely zero-shot fashion~\cite{herzig2018decoupling, chang2019zero}. Yet, such scenario requires acquiring the multi-domain dataset in the first place, 
and there is a significant gap between the accuracy of training with and without in-domain data~\cite{DBLP:journals/corr/abs-1809-08887}.
Our approach instead is able to synthesize data for the new domain, so the model is exposed to in-domain data while retaining the zero-shot property of no human-annotated data.


\paragraph{Pre-trained Models for Data Augmentation.}
Previous work showed that pre-trained models are very effective at generalizing natural language knowledge in a zero- and few-shot fashion~\cite{radford2019language, brown2020language}.
These models have been used to expand training data for various NLP classification tasks, by fine-tuning the model on a small seed dataset, then using conditioning on the class label to generate more data~\cite{anabytavor2019data, kumar2020data}.
\citet{kobayashi-2018-contextual} proposed using a bidirectional LSTM-based language model to substitute words that fit the context, conditioning on the class label to prevent augmentation from changing the class label. \citet{Wu_2019} used BERT~\cite{devlin2018bert} in a similar way, and \citet{hu2019learning} improved upon it by jointly fine-tuning BERT and the classifier. ~\citet{Semnani2019DomainSpecificQA} explored data augmentation for domain transfer using BERT.

These approaches rely on an initial dataset with many examples in each class, and therefore are not suitable for semantic parsing, where each logical form has only a few or even just one example. 




\paragraph{Neural Paraphrasing for Data Augmentation.}
The performance of many NLP tasks can be improved by adding automatically generated paraphrases to their training set. The general approach is to build a paraphrase generation model, usually a neural model (\citealp{prakash-etal-2016-neural}, \citealp{controlled-paraphrase}, \citealp{gupta2017deep}), using general-purpose datasets of paraphrase sentence pairs.

Data augmentation through neural paraphrasing models has been applied to various tasks such as sentiment analysis~\cite{controlled-paraphrase}, intent classification~\cite{roy-grangier-2019-unsupervised}, and span-based question answering~\cite{qanet}. Paraphrasing models may generate training examples that do not match the original label. Noisy heuristics, such as those employed by \newcite{qanet}, are not enough for semantic parsing, where paraphrases need to be semantically equivalent in a very strict and domain-dependent sense.
We propose a novel filtering approach, and show its effectiveness in reducing the noise of neural paraphrasing.

%% file: local-paraphrase.tex
\begin{table*}
\fontsize{8.5}{10}\selectfont
\setlength{\tabcolsep}{3pt}
\small
\centering
\begin{tabular}{llll}
\toprule
{\bf POS} & {\bf Annotation} & {\bf Example template} & {\bf Example utterance} \\
\midrule
is-a noun     & alumni of \ttvalue            & \tttable that$\vert$which$\vert$who is$\vert$are [noun phrase] \ttvalue   & people who are alumni of Stanford \\ 
has-a noun    & a \ttvalue degree             & \tttable with (a$\vert$an$\vert$the) \ttvalue [noun phrase]                                      & people with a Stanford degree\\
active verb   & graduated from \ttvalue       & \tttable that$\vert$which$\vert$who [verb phrase] \ttvalue                & people who graduated from Stanford \\
passive verb  & educated at \ttvalue          & \tttable [passive verb phrase] \ttvalue                                   & people educated at Stanford \\
adjective     & \ttvalue                      & \ttvalue\ \tttable                                                        & Stanford people \\
prepositional & from \ttvalue                 & \tttable [prepositional phrase] \ttvalue                                  & people from Stanford \\
 
\bottomrule
\end{tabular}
\caption{Annotations for ``alumniOf'' attribute with example templates and utterances in six POS categories, where
\tttable and \ttvalue denote the placeholders for table canonical annotations and values, respectively.}
\label{table:annotations}
\end{table*}

\section{\webqa Data Synthesis Pipeline}
\label{sec:schema2qa}

\system is based on \webqa \cite{xu2020schema2qa}, the state-of-the-art pipeline to generate high-quality training data for database QA at a low cost. 
\webqa first synthesizes utterance and formal representation pairs with a template-based algorithm, and then paraphrases utterances via crowdsourcing. 
The semantic parser is trained with both synthetic and paraphrased data, and tested on crowdsourced, manually annotated real questions.  

Instead of relying on crowdworkers to paraphrase and create variety from the synthesized canonical questions, \SQA uses a comprehensive set of 800 domain-independent templates, along with a few manual annotations for each attribute in each domain, to synthesize high-quality data. About 2\% of the synthesized data are manually paraphrased.  

Our previous work~\cite{xu2020schema2qa} shows that a parser trained on such dataset achieves 70\% accuracy on natural complex questions.
Table~\ref{table:thingtalk} shows a few questions that \SQA can parse and their representation in ThingTalk, which is a query language designed to support translation from natural language. 

\SQA answers long-tail questions well because its synthesized data have good coverage of possible questions asked, while showing great linguistic variety.  It synthesizes questions using generic question templates, which have place-holders to be substituted with domain-specific annotations that match the expected part-of-speech (POS) type. Table~\ref{table:annotations} shows how annotations of the 6 POS categories for the ``AlumniOf'' attribute are used in the example templates to synthesize example utterances. 
In total, six POS categories are identified: \textit{active verb} phrase, \textit{passive verb} phrase, \textit{adjective} phrase, \textit{prepositional} phrase, and two noun phrases: \textit{is-a noun} phrase which describes what the subject is, \textit{has-a noun} phrase which describes what the subject has. 
There is a wide variety in annotations for an attribute, and often only a subset of POS types is relevant to an attribute. 
It is thus challenging, often requiring multiple rounds of error analysis, to come up with these different annotations manually.

\section{Automatic Annotation}
\label{sec:auto-annotation}
Our \system toolkit 
automatically provides unambiguous attribute annotations for all parts of speech,
with the help of a neural paraphrasing model.

\subsection{Canonical Annotation}
AutoQA first derives a {\em canonical} annotation for each table and its attributes.  Where necessary, it splits the attribute name into multiple words (e.g. ``alumniOf'' turns into ``alumni of''). It then uses a POS tagger to identify the category of the canonical annotation. 

The canonical annotation is used both for training and as the starting point to identify alternative phrases for each attribute, hence it must be meaningful and unambiguous. When applying \system to an existing ontology, 
developers can override the table or attribute names if they are not meaningful or they are ambiguous.

\subsection{POS-based Annotation Extraction}
As shown in Table~\ref{table:annotations}, an attribute can be described in various ways in different parts of speech. It is not enough to retrieve synonyms of the canonical annotation, as all synonyms will have the same POS. Some synonyms may also be inappropriate for the domain, if generated without context. 

Our goal is to automatically derive all the other POS annotations given a canonical annotation. 
For example, the canonical annotation for the ``alumniOf'' attribute is ``alumni of {\em value}'' of POS ``is-a-noun'', 
as shown in the first row of Table~\ref{table:annotations}.
We wish to derive other ``is-a-noun'' annotations, as well as those in other POS categories in the table. 

Our solution is to synthesize questions using the templates for the POS of the canonical annotation, get paraphrases from a neural model, parse the paraphrases using the templates as grammar rules, and turn successful parses into annotations. 


\system first generates short example sentences for each attribute using its canonical annotation. 
We generate questions that ask for objects with a given value of the attribute, using the grammar templates for the POS of the canonical annotation for the attribute. 
We generate up to 10 sentences for each alternative in the grammar template, using a different value for each one. 

Second, \system obtains paraphrases for the generated sentences using a neural paraphraser based on the BART sequence-to-sequence model (Section \ref{sec:implementation}). 
To get more diverse paraphrases, we run 3 rounds of paraphrasing, where in each round we paraphrase the output of the previous round.  All the words are tagged with their POS. 
For example, with ``people who are alumni of Stanford'' as an input, we can get paraphrases such as 
``people with a Stanford degree'', as shown in the last column of Table~\ref{table:annotations}.

Third, \system parses the paraphrases using the templates (third column in Table~\ref{table:annotations}) as grammar rules. A phrase is considered a successful parse only if the ``table'' and the ``value'' match exactly and the POS of all placeholders match that of the corresponding words.  Correctly parsed phrases are then turned into annotations. 

Note that we generate only sentences that map to {\em selection} operations, such as ``show me people who are alumni of Stanford''. Selection questions include a sample value, ``Stanford'', for the attribute, which is useful to provide a better context for the paraphraser.  The paraphraser can generate phrases like ``find people from Stanford'', which is trivial to parse correctly.
In contrast, values are missing in {\em projection} questions, such as ``what institution are the people alumni of'', which makes paraphrasing and subsequent parsing harder. While we only paraphrase selection questions, the annotations identified will be used for all types of questions.

\subsection{Resolving Conflicts}

Neural paraphrasing is imperfect and can generate incorrect annotations. 
Our priority is to eliminate ambiguity: we do not worry as much about including 
nonsensical sentences in the training, as such sentences are unlikely to appear at test time. 
Consider a movie domain with both ``director'' and ``creator'' attributes. 
The paraphrasing model might generate the annotation ``creator'' for ``director''. 
To avoid generating such conflicted annotations within the domain, 
we detect annotations that appear in two or more attributes of the same type in the database. 
If such an annotation shares the same stem as one attribute name, it is assigned uniquely to that attribute.  
Otherwise, it is dropped entirely. As we train with data that is synthesized compositionally, we would rather lose a bit of variety than risk introducing ambiguity.



%% file: global-paraphrase.tex
\section{Automatic Paraphrasing}
\label{sec:autoparaphase}

Synthetic training data is good for providing coverage with a large number of perfectly annotated sentences, and to teach the neural semantic parser compositionality. However, grammar-based synthesis often results in clunky sentences and grammatical errors. 
In addition, even with 800 generic templates, the synthesized sentences still lack naturalness and variety.
In particular, people often compress multiple concepts into simpler constructions (sublexical compositionality~\cite{overnight}), e.g. ``books with at least 1 award'' can be simplified to ``award-winning books''.

Capturing these linguistic phenomena in the training data is not possible with a finite set of templates.
This is why paraphrasing is critical when training semantic parsers. Here we describe how we approximate manual paraphrases with a neural paraphrasing model. 



\subsection{Noise in Neural Paraphrasing}

Using automatically generated paraphrases for training is challenging. 
First, paraphrasing models output noisy sentences, partially due to the noise in the existing paraphrasing datasets\footnote{Most large-scale paraphrasing datasets are built using bilingual text~\cite{ppdb} and machine translation~\cite{mallinson-etal-2017-paraphrasing} or obtained with noisy heuristics~\cite{prakash-etal-2016-neural}. Based on human judgement, even some of the better paraphrasing datasets score only 68\%-84\% on semantic similarity (\citealp{parabank2}, \citealp{yang-etal-2019-end-end}).}.
We cannot accept paraphrases that change the meaning of the original sentence, which is represented by the logical form annotation. This noise problem exists even in human paraphrasing;   
\citet{overnight} reports that 17\% of the human paraphrases they collected changed the logical form.
Second, there is an inherent diversity-noise trade-off when using automatic generation. The more diverse we want to make the outputs, the noisier the model's output will be.
Third, the auto-paraphraser is fed with synthetic sentences, which have a different distribution compared to the paraphrase training set.

We have empirically found the following ways in which noise is manifested:
\begin{itemize}
    \item The output is ungrammatical or meaningless.
    \item The output changes in meaning to a different but valid logical form, or rare words like numbers and proper nouns are changed.
    \item The model is ``distracted'' by  the input sentence due to limited world knowledge.
``I'm looking for the book the dark forest'', is very different from ``I'm looking for the book \textit{in} the dark forest''.
    \item The model outputs sentence pairs that can be used interchangeably in general, but not in the specific application.
For example, ``restaurants close to my home'' and ``restaurants near me'' have different target logical forms.
    \item Automatically-generated annotations are not reviewed by a human to ensure their correctness. An example is the word ``grade'' instead of ``stars'' in the hotels domain.  Further paraphrasing these noisy sentences amplifies the noise. 
    \end{itemize}

\subsection{Paraphrase Filtering}
How do we produce semantically correct paraphrases and yet obtain enough variety to boost the accuracy of the parser?
Our approach is to generate high variety, and then filter out noisy sentences.
More specifically, we feed auto-paraphrased sentences to a parser trained on only synthetic sentences. We accept the sentences as correct paraphrases only if this parser outputs a logical form equal to the original logical form.

Correct paraphrases are then used to train another parser from scratch, which will have a higher accuracy on the natural validation and test sets. The first parser can correctly parse the examples present in the synthetic set, e.g. ``I am looking for the movies which have Tom Hanks in their actors with the largest count of actors.''. It also generalizes to paraphrased sentences like ``I'm looking for Tom Hanks movies with the most actors in them.''. Paraphrased sentences like this are added to the training set to generate a second parser. 
This second parser can generalize to an even more natural sentence like ``What is the Tom Hanks movie with the biggest cast?''
This iterative process, as shown in Fig.~\ref{fig:diagram}, can be repeated multiple times. 

This idea is borrowed from self-training~\cite{mcclosky-2006, he2019revisiting}, where a model is used to label additional unlabeled data. Self-training requires an initial \emph{good-enough} model to label data with, and optionally a filtering mechanism that is more likely to remove incorrect labels than correct labels~\cite{yarowsky-1995-unsupervised}.
We use a parser trained on a synthetic dataset as our initial \emph{good-enough} model. The following two observations are the intuition behind this decision:

\begin{enumerate}
    \item Paraphrases of a synthetic dataset are still relatively similar to that set. Thus, a parser trained on synthetic data, which delivers near perfect accuracy for the synthetic data, has a very high accuracy on the paraphrased data as well.
    
    \item Unlike classification tasks, the set of valid logical forms in semantic parsing is so large that outputting the right logical form by chance is very unlikely.
\end{enumerate}

Note that this filtering scheme might throw away a portion of correct paraphrases as well, but filtering out noisy examples is more important. The second observation ensures that the number of false positives is low.

\subsection{Coupling Auto-Annotator with Auto-Paraphraser}

Since both auto-annotation and auto-paraphrasing use a neural paraphraser, here we contrast them and show how they complement each other.  

Auto-annotation provides alternative expressions with different POS for a single attribute at a time.
The input sentences are simpler, so paraphrases are more likely to be correct, and they are filtered if they cannot be parsed correctly with the grammar rules.
This makes it easier to coax more diverse expressions on the attribute from the paraphraser without having to worry about noisy outputs. 

Annotations extracted by the auto-annotator are amplified as the synthesizer uses them to compose many full sentences, which are used to train the first parser with sufficient accuracy for self-training.  

The auto-paraphraser, on the other hand, is applied on all synthesized data. It not only produces more natural alternative phrases for complex sentences, but also generates domain-specific and value-specific terminology and constructs. These two tasks complement each other, as supported by the empirical results in Section \ref{sec:ablation}.


%% file: experiments.tex
\section{Experiments}
\label{sec:experiments}

\begin{table*}[htb]
\setlength{\tabcolsep}{5pt}
\renewcommand{\arraystretch}{0.8}
\small
\centering
\begin{tabular}{lllrrrrrrr}
\toprule
& & \multicolumn{2}{r}{\bf Restaurants} & {\bf People} & {\bf Movies} & {\bf Books} & {\bf Music} & {\bf Hotels} & {\bf Average}\\
\midrule
\multicolumn{3}{l}{\# Attributes} & 25 & 13 & 16 & 15 & 19 & 18 & 17.7\\
\midrule
\multirow{6}{*}{Train} & \multirow{3}{*}{\webqa} 
& {\# of Annotations} & 122 & 95 & 111 & 96 & 103 & 83 & 101.7\\
&& {Synthesized Data} & 270,081 & 270,081 & 270,081 & 270,081 & 270,081 & 270,081 & 270,081 \\
&&{Human Paraphrase} & 6,419 & 7,108 & 3,774 & 3,941 & 3,626 & 3,311 & 4,697\\
\cmidrule{2-10}
& \multirow{3}{*}{\system} 
& {\# of Annotations} & 151 & 121 & 157 & 150 & 144 & 160 & 147.2\\
&& {Synthesized Data} & 270,081 & 270,081 & 270,081 & 270,081 & 270,081 & 270,081 & 270,081 \\
&&{Auto Paraphrase} & 280,542 & 299,327 & 331,155 & 212,274 & 340,721 & 285,324 & 291,557 \\
\midrule
Dev  &&&  528  & 499 & 389 & 362 & 326 & 443 & 424.5\\
\midrule
Test &&&  524  & 500 & 413 & 410 & 288 & 528 & 443.8\\
%
%
\bottomrule
\end{tabular}
\caption{Size of \SQA and \system datasets}
\label{table:sqa-dataset}
\end{table*}

In this section, we evaluate the effectiveness of our methodology: can a semantic parser created with \system approach the performance of human-written annotations and paraphrases? 
We evaluate on two different benchmark datasets: the \SQA dataset~\cite{xu2020schema2qa} and the \overnight dataset~\cite{overnight}.
\subsection{\system Implementation}
\label{sec:implementation}

\paragraph{Paraphrasing Model.}
We formulate paraphrasing as a sequence-to-sequence problem and use the pre-trained BART large model~\cite{bart}. BART is a Transformer~\cite{vaswani2017attention} neural network trained on a large unlabeled corpus with a sentence reconstruction loss. We fine-tune it for 4 epochs on sentence pairs from \textsc{ParaBank 2}\xspace~\cite{parabank2}, which is a paraphrase dataset constructed by back-translating the Czech portion of an English-Czech parallel corpus. We use a subset of 5 million sentence pairs with the highest dual conditional cross-entropy score \cite{dual-conditional-cross-entropy}, and use only one of the five paraphrases provided for each sentence.
We experimented with larger subsets of the dataset and found no significant difference.
We use token-level cross-entropy loss calculated using the gold paraphrase sentence. To ensure the output of the model is grammatical, during training, we use the back-translated Czech sentence as the input and the human-written English phrase as the output. Training is done with mini-batches of 1280 examples where each mini-batch consists of sentences with similar lengths\footnote{This reduces the number of pad tokens needed, and makes training faster.}.

We use nucleus sampling~\cite{holtzman2019curious} with top-$p$$=$0.9 and generate 5 paraphrases per sentence in each round of paraphrasing. We use greedy decoding and 4 temperatures~\cite{Ficler_2017} of 0.3, 0.5, 0.7 and 1.0  to generate these paraphrases. Note that the input dataset to each paraphrasing round is the output of the previous round, and we have one round for \webqa and three rounds for \overnight experiments.


\paragraph{Semantic Parsing Model.}
We adopt our previously proposed BERT-LSTM model~\cite{xu2020schema2qa} as the semantic parsing model.
The model is a sequence-to-sequence neural network that uses a BERT pre-trained encoder~\cite{devlin2018bert}, coupled with an LSTM decoder~\cite{hochreiter1997long} with attention~\cite{bahdanau2014neural}. The model uses a pointer-generator decoder~\cite{see2017get} to better generalize to entities not seen during training. The model was implemented using the Huggingface Transformers library~\cite{Wolf2019HuggingFacesTS}.
We use the same hyperparameters as \newcite{xu2020schema2qa} for all experiments. The model has approximately 128M parameters.




\begin{table*}[htb]
\renewcommand{\arraystretch}{0.9}
\setlength{\tabcolsep}{4pt}
\small
\centering
\begin{tabular}{lccccccc}
\toprule
Model & {\bf Restaurants} & {\bf People} & {\bf Movies} & {\bf Books} & {\bf Music} & {\bf Hotels} & {\bf Average}\\
\midrule 
\webqa~\cite{xu2020schema2qa} &
    69.7 & 
    75.2 & 
    70.0 & 
    70.0 & 
    63.9 & 
    67.0 &
    69.3 \\
\webqa w/o manual annotation \& paraphrase & 
   30.0 & 
   30.4 & 
   36.6 & 
   34.9 & 
   33.7 & 
   59.7 &
   37.6\\
\system & 
   65.3 & 
   64.6 & 
   66.1 & 
   54.1 & 
   57.3 &
   70.1 &
   62.9\\

\bottomrule
\end{tabular}
\caption{Test accuracy of \system on the \SQA dataset. For the hotel domain, \newcite{xu2020schema2qa} only report transfer learning accuracy, so we rerun the training with manual annotations and human paraphrases to obtain the accuracy for hotel questions.}
\label{tab:schemaorg_test}
\end{table*}



\begin{table*}[htb]
\renewcommand{\arraystretch}{0.9}
\setlength{\tabcolsep}{4pt}
\small
\centering
\begin{tabular}{lccccccc}
\toprule
& {\bf Restaurants} & {\bf People} & {\bf Movies} & {\bf Books} & {\bf Music} & {\bf Hotels} & {\bf Average}\\
\midrule
\webqa~\cite{xu2020schema2qa} & 70.8 & 74.9 & 75.3 & 80.7 & 71.8 & 69.3 & 73.8\\
\midrule
\webqa (w/o manual annotation \& paraphrase) & 33.9 & 32.7 & 35.7 & 39.9 & 37.1  & 61.6 & 40.2   \\
\midrule
\system & 69.5 & 66.1 & 68.0 & 67.6 & 66.9 & 66.6 & 67.4\\
-- Auto-annotation & 43.2 & 50.1 & 51.4 &  59.6 & 49.7 & 67.3 & 53.5 \\
-- Auto-paraphrase &  62.1 & 50.5 & 62.7 & 61.5 & 58.6 & 59.1 & 59.1 \\
-- Paraphrase filtering & 50.4 & 48.0 & 55.0 & 44.1 & 53.5 & 44.7 & 49.3 \\
\bottomrule
\end{tabular}
\caption{Ablation study on Schema2QA development sets. Each ``--'' line removes only that feature from AutoQA.}
\label{tab:schemaorg_dev}
\end{table*}

\subsection{Applying \system to \webqa}
\label{sec:experiments-schema2qa}
We first apply \system to the \webqa dataset, a semantic parsing dataset that targets the ThingTalk query language, and uses Schema.org as the database schema. Queries are performed against structured data crawled from websites in 6 domains: restaurants (using data from Yelp), people (from LinkedIn), hotels (from the Hyatt hotel chain), books (from Goodreads), movies (from IMDb), and music (from Last.fm).

The \SQA training data set was created using synthesis based on manual field annotations and human paraphrasing, while its evaluation data was crowdsourced by showing the list of attributes to workers and asking them for natural questions. 
The evaluation data contains complex questions referring up to 6 attributes, with comparisons and relational algebra operators: join, selection, projection, sort, and aggregates.

In our experiments, we use the 
\SQA validation and test sets, but {\em not} the training data.
We synthesize our own training data using the same 800 templates, and replace the manual annotations with our auto-annotation and the manual paraphrases with auto-paraphrases. 

For auto-annotation to work, the table and attribute names must be meaningful and unambiguous as discussed in Section~\ref{sec:auto-annotation}.  
We found it necessary to override the original names in only three cases. 
In the restaurants domain, ``starRating'' is renamed to ``michelinStar'' to avoid ambiguity with ``aggregateRating''.
In the people domain, ``addressLocality'' is renamed to ``homeLocation'' to avoid confusion with ``workLocation''.
In the music domain, ``musicRecording'' is renamed to ``song'' to better match natural language. 

When applying auto-paraphrasing, we preprocess the questions to replace entity placeholders (e.g. TIME\_0) with an equivalent token in natural language (e.g. 2pm), then postprocess the outputs to restore them. This way, the neural network does not have to deal with these tokens which it has not seen during its pre-training.

As shown in Table~\ref{table:sqa-dataset}, \system generates about 45\% more attribute annotations, and produces 60 times larger paraphrase sets, compared with the original \webqa training set. Although \system's training set is larger than \webqa's, we note that in our experiments, adding more synthetic data to \webqa did not improve its accuracy any further.
We compare the diversity of the two datasets using distinct-1 and distinct-2 metrics~\cite{li2016diversity} which measure the ratio of distinct unigram and bigrams in the datasets. \system's training sets have about 35\% higher distinct-1 and 60\% higher distinct-2.

\subsubsection{Evaluation}
Our evaluation metric is \emph{logical form accuracy}: the logical form produced by our parser must exactly match the one in the test set. As shown in Table~\ref{tab:schemaorg_test}, 
\system achieves an average accuracy of 62.9\% in six domains, only 6.4\% lower
compared to the models trained with manual attribute annotations and human paraphrases.
The difference is mainly because paraphraser fails to generate a few common phrases in some cases. 
For example, it fails derive ``employee'' or ``employed by'' from the canonical annotation ``works for'', which is 
quite common in the evaluation set.
Compared with the baseline models trained with data generated by \webqa but without manual annotation and human paraphrase,
\system improves the accuracy by 25.3\%.
This result is obtained on naturally sourced test data, as opposed to paraphrases. 
This shows that \system is effective for bootstrapping question answering systems for new domains, without any manual effort in creating or collecting training data. 



\begin{table*}[htb]
\fontsize{8}{10}\selectfont
\setlength{\tabcolsep}{1pt}
\centering
\scalebox{1}{\begin{tabular}{p{0.25\textwidth}|P{0.035\textwidth}P{0.035\textwidth}|P{0.035\textwidth}P{0.035\textwidth}|P{0.035\textwidth}P{0.035\textwidth}|P{0.035\textwidth}P{0.035\textwidth}|P{0.035\textwidth}P{0.035\textwidth}|P{0.035\textwidth}P{0.035\textwidth}|P{0.035\textwidth}P{0.035\textwidth}|P{0.035\textwidth}P{0.035\textwidth}|P{0.035\textwidth}P{0.035\textwidth}}
\toprule
    \multicolumn{1}{l}{Model} & 
    \multicolumn{2}{c}{Basketball} & 
    \multicolumn{2}{c}{Blocks} & 
    \multicolumn{2}{c}{Calendar} & 
    \multicolumn{2}{c}{Housing} & 
    \multicolumn{2}{c}{Publications} & 
    \multicolumn{2}{c}{Recipes} & 
    \multicolumn{2}{c}{Restaurants} & 
    \multicolumn{2}{c}{Social} &  
    \multicolumn{2}{c}{Average} \\
\hline
\bf Only in-domain human data &&&&&&&&&&&&&&&&& \\
\citet{cao2019semantic} & 
    - & 88.0 &
    - & 65.2 &
    - & 80.7 &
    - & 76.7 &
    - & 80.7 &
    - & 82.4 &
    - & 84.0 &
    - & 83.8 &
    - & 80.2 \\
\citet{chen-etal-2018-sequence} & 
    - & 88.2 &
    - & 61.4 &
    - & 81.5 &
    - & 74.1 &
    - & 80.7 &
    - & 82.9 &
    - & 80.7 &
    - & 82.1 &
    - & 79.0 \\
\citet{Damonte_2019} &
   69.6 & - &
   25.1 & - &
   43.5 & - &
   29.6 & - &
   32.9 & - &
   58.3 & - &
   37.3 & - &
   51.2 & - &
   43.4 & - \\
BERT-LSTM & 
    84.1& 87.5 & 
    42.6& 62.4 & 
    58.3& 79.8 & 
    48.7& 70.4 & 
    64.6& 76.4 & 
    68.5& 75.9 & 
    55.4& 82.8 & 
    70.4& 81.9 & 
    61.6& 75.0 \\
\hline
\bf Only out-of-domain human data &&&&&&&&&&&&&&&&& \\
\citet{decoupling2018} & 
    - & - &
    - & 28.3 &
    - & 53.6 & 
    - & 52.4 &
    - & 55.3 & 
    - & 60.2 &
    - & 61.7 &
    - & 62.4 &
    - & 53.4 \\
\hline
\bf No human data &&&&&&&&&&&&&&&&& \\
\citet{marzoev2020unnatural} &
    47 & - &
    27 & - &
    32 & - &
    36 & - &
    34 & - &
    49 & - &
    43 & - &
    28 & - &
    37 & - \\
BERT-LSTM (Synthetic only) &  
    29.7&  31.5&  
    27.6& 37.8 &  
    28.0& 34.5&  
    18.0& 32.8&  
    28.0& 37.3&  
    40.7& 48.6&  
    34.9& 47.0&  
    16.1& 24.2&  
    27.9& 36.7\\
BERT-LSTM w/ \system (ours) &
    70.1&  73.9& 
    38.4&  54.9& 
    58.9&  72.6& 
    51.9&  70.9& 
    56.5& 74.5& 
    64.4& 68.1& 
    57.5&  78.6& 
    47.2&  61.5& 
    55.6&    69.8\\
\bottomrule
\end{tabular}}
\caption{Logical form accuracy (left) and answer accuracy (right) percentage on the \overnight test set. Numbers are copied from the cited papers. We report the numbers for the BL-Att model of \citet{Damonte_2019}, Att+Dual+$\mathcal{LF}$ of \citet{cao2019semantic}, {\textsc{ZeroShot} model}\xspace of \citet{decoupling2018}, and the Projection model of \citet{marzoev2020unnatural}. \citet{decoupling2018} do not evaluate on the Basketball domain.}
\label{table:overnight}
\end{table*}

\subsubsection{Ablation Study}
\label{sec:ablation}

We conduct an ablation study on the development set to evaluate how each part of our methodology contributes to the accuracy.
We subtract different components from \system, generate the training data, 
and run the experiment with the same hyperparameters. When paraphrase filtering is removed, we still use simple string matching to remove erroneous paraphrases where entities and numbers in the utterance do not match the logical form.

As shown in Table~\ref{tab:schemaorg_dev}, \system reaches an overall accuracy of 67.4\%, 6.4\% lower than models trained with human annotations and human paraphrases. \system outperforms the baseline trained on synthetic data generated from the canonical annotation by 27.2\%. 
This indicates that \system is an efficient and cost-effective replacement for manual annotation and paraphrasing.

On average, applying only auto-paraphrase on synthetic data based on canonical annotations without auto-annotation
achieves 53.5\%, which is 13.9\% lower than the full \system.
Applying only auto-annotation without auto-paraphrase obtains 59.1\%, and is 8.3\% lower than \system.
This shows that the two components of \system complement each other to achieve the best performance.

If auto-paraphrase is used without filtering, not only does it not improve the accuracy, but also the average accuracy drops by 18\%. This shows that without filtering, even a paraphraser with a large pre-trained neural model like BART cannot be used for semantic parsing due to noisy outputs.

\subsection{Applying \system to \overnight}
\label{sec:overnight}

To evaluate if the \system methodology generalizes to different types of databases, logical forms, 
and templates, we apply \system on the well-known \overnight benchmark. 
\overnight is a semantic parsing dataset with questions over a knowledge base with very few entities across 8 domains. The dataset was constructed using paraphrasing; both training and test sets are paraphrased from the same set of synthetic sentences. 

We train the BERT-LSTM model on data synthesized from \overnight templates with both auto-annotation and auto-paraphrase. 
Auto-annotation is limited to two parts of speech, since \overnight 
uses a very simple template set to synthesize training examples, with only placeholders for active verb phrase and noun phrase. 
We use the standard train/test split and following previous work, use 20\% of the human paraphrases from the original training set for validation, so that validation and test sets are from the same distribution.  


We evaluate both \emph{logical form accuracy} and \emph{answer accuracy}, which checks whether the answer retrieved from the knowledge base matches the gold answer. The model outputs a ranked list of logical forms for each input question using beam search with 25 beams, and chooses the first output that is syntactically valid. Other than this, all models and hyperparameters are the same as Section~\ref{sec:experiments}.

In Table~\ref{table:overnight}, we compare our technique to other approaches that do not use in-domain human data. 
They are either synthetic-only \cite{marzoev2020unnatural} or use human data from other \overnight domains \cite{decoupling2018}. For reference, we also include two of the best-performing models that use in-domain human data~\cite{cao2019semantic, chen-etal-2018-sequence}\footnote{These are the best-performing models among those that use training data from a single domain, and do not do transfer-learning from other domains or datasets.}. 

Whereas \webqa dataset has naturally sourced evaluation and test data, 
\overnight evaluates on human paraphrase data.
Evaluating with paraphrase data is not as meaningful, and makes the benchmark easier for models trained with human paraphrase data~\cite{geniepldi19}.   
Nonetheless, \system achieves an average logical form accuracy of 55.6\% and answer accuracy of 69.8\%, which is only 5.2\% lower than the same parser trained with human paraphrases, and matches its performance in the housing domain.
Compared to other zero-shot models trained with no in-domain data,
\system outperforms the state of the art by 18.6\% and 16.4\% on logical form accuracy and answer accuracy, respectively.
This shows that by generating diverse and natural paraphrases in domain, 
\system can reach comparable performance with models with human training data, 
and is much more accurate compared to other zero-shot approaches. 

%% file: conclusion.tex
\section{Discussion}
In this work, we propose \system, a methodology and a toolkit to automatically
create a semantic parser given a database. 
We test \system on two different datasets with different target logical forms
and data synthesis templates.
On both datasets, \system achieves comparable accuracy to state-of-the-art QA systems
trained with manual attribute annotation and human paraphrases.


\system relies on a neural paraphraser trained with an out-of-domain
dataset to generate training data.
We suspect the methodology to be less effective for domains full of jargon.  
Even for common domains, \system sometimes failed to generate some common phrases. 
Further improvement on neural paraphraser is needed to generate more diverse outputs. 
Future work is also needed to handle attributes containing long free-form text, as \system currently
only supports database operations without reading comprehension.

%% file: appendix.tex
\appendix
\section{The Cost of \system}
\label{sec:appendix}

The only form of cost in \system's methodology is compute cost. Here we mention more details with regards to that. 
To use \system for a new domain, the following steps will have to be executed to generate the final training set. Numbers are for the \webqa dataset, and batch sizes are set to maximize GPU utilization. For steps that do not need GPU we use AWS m5.4xlarge machines (16 vCPU and 64 GiB of memory). For GPU we use AWS p3.2xlarge machines (16GB V100 GPU, 8vCPUs, 61 GiB of memory).

\begin{itemize}
    \item Automatic annotation: This step runs inference using the BART paraphraser model as introduced in Section~\ref{sec:implementation}, it takes less than 10 minutes on CPU for each domain. 
    
    \item Template-based data synthesizer: This step synthesize data with annotation generated by auto-annotator. Depending on the domain, it takes between 3 to 5 hours on a CPU machine. 
    
    \item Training a parser with the synthetic dataset to use as filter: We train the BERT-LSTM model for 4000 iterations only, as we empirically observed that training more than that does not improve the quality of the filter. This takes less than half an hour on a single-GPU machine.
    
    \item Automatic paraphrasing and filtering: This step uses the fine-tuned BART large model, which has about 400M parameters, to generate 5 paraphrases per input, and then the BERT-LSTM parser, which has 128M parameters, to filter those paraphrases. Note that no training is done in this step. In our experiments, this step takes less than 3 GPU-hours.
    
    \item Training of the semantic parser: Similar to training the filter, but we train for 60000 iterations, and it takes less than 6 GPU-hours.
\end{itemize}

Given these numbers, the approximate total cost to get a semantic parser for one \webqa domain using Amazon Web Services is \$33.
